\documentclass{article}
\usepackage{spconfa4}

\usepackage{epsfig}
\usepackage{graphicx}
\usepackage{amsmath}
\usepackage{amssymb}
\usepackage{upgreek}
\usepackage{appendix}
\usepackage[ruled]{algorithm2e}
\usepackage{color}
\usepackage{frame}
\usepackage{multirow}

\graphicspath{{figures/}}
\newcommand{\tabincell}[2]{\begin{tabular}{@{}#1@{}}#2\end{tabular}}

\definecolor{iccvblue}{rgb}{0.21,0.49,0.74}
\usepackage[pagebackref,breaklinks,colorlinks,allcolors=iccvblue]{hyperref}

\usepackage{booktabs}
\usepackage{times}
\usepackage{diagbox}
\usepackage{cleveref}
\usepackage{subcaption}

\newcommand{\etal}{\textit{et al.}}

\begin{document}

\title{Decoupled Geometric Parameterization and its Application \\ in Deep Homography Estimation}



\name{
\begin{minipage}{0.8\textwidth}\centering\em
Yao~Huang$^{1}$, Si-Yuan~Cao$^{2}$, Yaqing~Ding$^{3}$, Hao~Yin$^{1}$, Shibin~Xie$^{1}$, Shuting~Wang$^{1}$, Zhijun~Fang$^{1}$, 
Jiachun~Wang$^{4}$, Shen~Cai$^{1,\ast}$, Junchi~Yan$^{5}$, and~Shuhan~Shen$^{6,7,\ast}$
\end{minipage}
}

\address{
$^{1}$ School of Computer Science and Technology, Donghua University\\
$^{2}$ College of Information Science and Electronic Engineering, Zhejiang University\\
$^{3}$ Visual Recognition Group, Czech Technical Univeresity in Prague\\
$^{4}$ Huawei Technologies\\
$^{5}$ School of Artificial Intelligence, Shanghai Jiao Tong
 University\\
$^{6}$ Institute of Automation, Chinese Academy of Sciences\\
$^{7}$ School of Artificial Intelligence, University of Chinese Academy of Sciences\\
$^{\ast}$ Corresponding authors: Shen Cai (hammer\_cai@163.com) and Shuhan Shen (shshen@nlpr.ia.ac.cn).
}

\maketitle

\begin{abstract}
Planar homography, with eight degrees of freedom (DOFs), is fundamental in numerous computer vision tasks. While the positional offsets of four corners are widely adopted (especially in neural network predictions), this parameterization lacks geometric interpretability and typically requires solving a linear system to compute the homography matrix. This paper presents a novel geometric parameterization of homographies, leveraging the similarity-kernel-similarity (SKS) decomposition for projective transformations. Two independent sets of four geometric parameters are decoupled: one for a similarity transformation and the other for the kernel transformation. Additionally, the geometric interpretation linearly relating the four kernel transformation parameters to angular offsets is derived. Our proposed parameterization allows for direct homography estimation through matrix multiplication, eliminating the need for solving a linear system, and achieves performance comparable to the four-corner positional offsets in deep homography estimation. 
\end{abstract}

\section{Introduction}
\label{sec:intro}

Planar homography, also known as two-dimensional (2D) projective transformation, is typically represented as a $3\!*\!3$ matrix. Due to homogeneous equality, a homography matrix contains 8 degrees of freedom (DOFs). Homography estimation is essential for many computer vision tasks, such as camera calibration~\cite{Zhang_PAMI00}, pose estimation~\cite{IPPE_IJCV14}, image stitching~\cite{ImageStitch}, and simultaneous localization and mapping (SLAM)~\cite{SLAM_1,SLAM_3}. Traditional homography estimation often involves the extraction and matching of key points between images, outlier removal based on the random sample consensus (RANSAC) scheme~\cite{RANSAC}, and the use of the direct linear transformation (DLT) algorithm~\cite{Hartley2003Multiple}[p.~88] for homography computation. Similarly, most deep homography estimation algorithms take two images as network inputs and predict point correspondences for the subsequent DLT solver, such as the positional offsets (displacements) of four corners~\cite{DHN16,Nowruzi2017,MHN,LocalTrans,IHN,UDHN,RHWF, RAL18,MCNet,HomoGAN} and eight coefficients for the optical flow bases~\cite{MotionBase, MotionBase_1}. 

\begin{figure}[t]
\begin{center}
\includegraphics[width=1.0\linewidth]{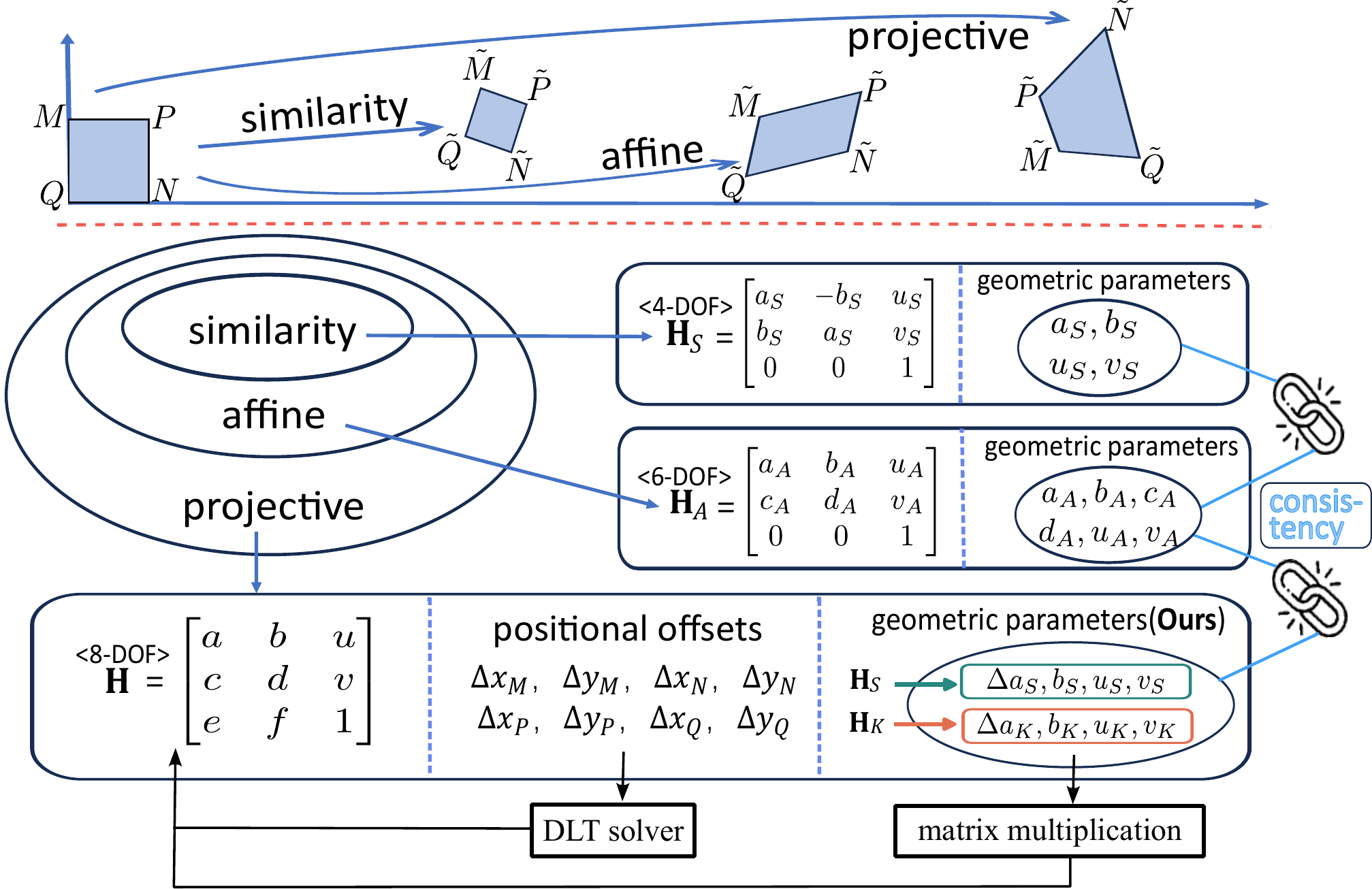}
\end{center}
\vspace{-5mm}
\caption{Parameterization in hierarchical 2D geometric transformations. Upper: shape distortion. Lower: matrix representation and parameterization. Our geometric parameterization is decoupled into two sets of four parameters: one for a similarity transformation $\mathbf{H}_S$ and the other for the kernel transformation $\mathbf{H}_K$.} 
\label{fig:transform_Hierarchy}
\vspace{-3mm}
\end{figure}

Among all parameterizations of homography, positional offsets (P.O.) of four corners are the most common in the task of deep homography estimation, from the pioneering work~\cite{DHN16} in 2016 to the most recent state-of-the-art (SOTA) works~\cite{RHWF,MCNet}. Homography parameterization based on P.O. has been thoroughly validated for effectiveness across various types of datasets and network architectures. This is explicable, as the displacements of corresponding points are explicit (visual) features between images, facilitating extraction by neural networks~\cite{SuperGlue,Patch2Pix,FlowNet,PWCNet}. However, after obtaining the four-corner P.O., previous deep homography estimation works still require the DLT algorithm to complete the homography computation by solving a system of linear equations. Moreover, the P.O. of corresponding points between images are implicit parameters for transformation and do not possess straightforward geometric meanings, such as scale and rotation in 2D space.

Hierarchical 2D geometric transformations~\cite{Hartley2003Multiple}[p.~44] and their parameterization are outlined in~\cref{fig:transform_Hierarchy}. A similarity transformation $\mathbf{H}_S$ has four geometric parameters $\{a_S, b_S, u_S, v_S\}$, with the last two representing 2D translation. The first two parameters can also be interpreted as the combination of isotropic scaling $s_S$ and planar rotation $\theta_S$, where $a_S\!=\!s_S\!*\!\cos(\theta_S)$ and $b_S\!=\!s_S\!*\!\sin(\theta_S)$. An affine transformation $\mathbf{H}_A$ introduces two additional parameters $\{c_A, d_A\}$, caused by the scaling along the other orthogonal direction and the shear transformation. Homography, relative to affine transformations, has two more parameters $\{e, f\}$ indicating projective components. In the context of similarity and affine transformation estimation, previous deep learning based methods predict geometric parameters either directly in the transformations or through their decomposition matrices~\cite{sim_1,affine_1,affine_2,affine_3}. Therefore, the current mainstream parameterization of homography using four-corner P.O. is inconsistent with the geometric parameters employed in similarity and affine estimation tasks, showing a significant discrepancy. In contrast, in tasks like pose estimation, most methods~\cite{Pose_param_1,Pose_param_2,Pose_param_3,Pose_param_4}  estimate geometric parameters of 3D translation and rotation (e.g., Euler angles, rotation axis and angle, or quaternions) rather than P.O. of corresponding points between images. This pose parameterization avoids inconsistencies in geometric representation and eliminates the need for a post-processing step using a P3P (Perspective-3-Point) or PnP (Perspective-n-Point) solver.

In this paper, we rethink the geometric parameterization of homography. For similarity transformations, we derive linear transformations between their geometric parameters and two-corner positional offsets. For projective transformations, leveraging a recent work~\cite{Mine_SKS-I} which proposes decomposing homography into similarity-kernel-similarity (SKS) sub-transformations, we propose an improved SKS adapting to deep homography estimation. Specifically, homography is decoupled into two independent sets of four geometric parameters: one set in a similarity transformation and the other in the kernel transformation within a series of sub-transformation chains. To demonstrate that the four geometric parameters of the kernel transformation are suitable for estimation, we establish a linear relationship between the geometric parameters of the kernel transformation and four angular offsets (A.O.), which are also explicit features of the image pair. The proposed homography parameterization has been validated for accuracy and robustness across multiple datasets and neural network architectures in the application of deep homography estimation.

The contributions of the paper are as follows:

- We analyze why neural networks can effectively estimate the geometric parameters of similarity and affine transformations, emphasizing the existence of linear transformations with P.O. of corners.

- We improve the SKS homography decomposition to suit the neural network estimation of eight geometric parameters, four in a similarity transformation and four in the kernel transformation.

- We introduce a geometric interpretation of angular offsets for the 4-DOF kernel transformation, decoupling the 8-DOF projective distortion into two independent sets: four P.O. and four A.O., existing in two point correspondences.

- Our proposed homography parameterization matches the performance of the standard four-corner P.O. across multiple datasets and neural networks, free of solving a linear system to obtain the homography.

\section{Related Works}
\label{sec:preliminaries}

\subsection{Traditional Homography Estimation}
Traditional homography estimation often involves interest point extraction and matching, such as hand-crafted SIFT~\cite{SIFT}, SURF~\cite{SURF2006}, ORB~\cite{ORB}, and deep learning based LIFT~\cite{LIFT_ECCV16}, SOSNet~\cite{SOSNet_CVPR19}, SuperPoint~\cite{SuperPoint18}, and SuperGLUE~\cite{SuperGlue}. Next, the DLT solver~\cite{Hartley2003Multiple} is applied under the random sample consensus (RANSAC)~\cite{RANSAC} framework or its variants (e.g., USAC~\cite{Raguram2013USAC} and MAGSAC~\cite{MAGSAC}) to estimate homography between two images with outliers.

\begin{figure*}[t]
\begin{center}
\includegraphics[width=\textwidth]{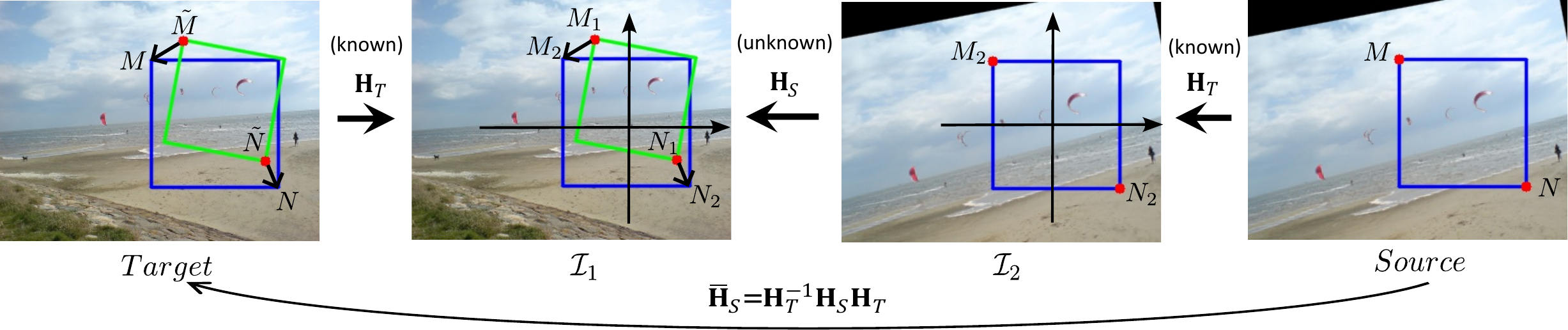}
\end{center}
\vspace{-5mm}
\caption{Our similarity estimation process utilizing translation normalization. Among the three sub-transformations, the translation transformation \(\mathbf{H}_{T}\), which normalizes the square's center, is already known. The similarity transformation \( \mathbf{H}_{S}\) serves to map $\{M_2, N_2\}$ to $\{M_1, N_1\}$, as well as to transform the normalized image $\mathcal{I}_2$ to $\mathcal{I}_1$.}
\label{fig:similarity_estimation}
\vspace{-2mm}
\end{figure*}
\subsection{Deep Homography Estimation}

[\textbf{Four-corner P.O.}] DeTone~\etal~\cite{DHN16} pioneered the integration of deep learning into homography estimation by predicting four-corner P.O. of a square sub-image. Subsequently, Nguyen~\etal~\cite{RAL18} introduced a photometric loss function to train the neural network in an unsupervised learning manner. Le~\etal~\cite{MHN} incorporated a dynamic mask network within the multi-scale framework to adapt to varying scenes. Zhang~\etal~\cite{UDHN, UDHN_pami} incorporated mask mechanisms to refine homography estimation by focusing on relevant image regions. Cao~\etal~\cite{IHN} introduced IHN, adopting the iterative strategy to improve estimation accuracy without additional parameters. Moreover, Cao~\etal~\cite{RHWF} further refined this model in RHWF, incorporating transformations and image deformations to enhance estimation capabilities. Zhu~\etal~\cite{MCNet} developed MCNet, which employs multi-scale correlation searches to optimize the efficiency and accuracy of homography estimation. All these approaches predict four-corner P.O. and require the DLT algorithm as an algebraic solver to compute homography in a post-processing step.

\vspace{1mm}
\noindent [\textbf{ICLK}] Inspired by the inverse compositional Lucas-Kanade (ICLK) algorithm~\cite{ICLK}, Chang~\etal~\cite{CLKN} introduced the cascaded Lucas-Kanade network (CLKN), which combined the parameterization of ICLK with deep learning in a cascade structure. Subsequently, Zhao~\etal~\cite{DLKN} developed the DLKFM network for multi-modal image alignment.

\vspace{1mm}
\noindent [\textbf{Motion Basis}] Both Liu~\etal~\cite{MotionBase_1} and Ye~\etal~\cite{MotionBase} proposed to represent homography using motion basis. By learning eight lower-dimensional motion basis, neural networks efficiently captured essential homography patterns.


\subsection{SKS Homography Decomposition}
\label{subsec:SKS}
\vspace{-1mm}

Cai~\etal~\cite{Mine_SKS-I} proposed a novel similarity-kernel-similarity (SKS) decomposition for $4$-point homography computation. A homography $\mathbf{H}$ is decomposed into three sub-transformations as
\begin{equation}
\mathbf{H}=\mathbf{H}_{S_2}^{-1}\mathbf{H}_K\mathbf{H}_{S_1},
\label{equ:HS}
\end{equation}
where $\mathbf{H}_{S_1}$ and $\mathbf{H}_{S_2}$ are similarity transformations, mapping two points in the source plane and their correspondences in the target plane to the normalized points $[\mp1,0]^\top$, respectively. The kernel transformation $\mathbf{H}_{K}$ is expressed as
\begin{equation}
\mathbf{H}_K =
\begin{bmatrix}
a_K & u_K & b_K \\
     & 1 &   \\
 b_K & v_K & a_K \end{bmatrix},
\label{equ:H_K}
\end{equation}
where the 4-DOF $\mathbf{H}_{K}$ is calculated from the other two correspondences and imposes the projective distortion between the two similarity-normalized planes. 

Although the SKS decomposition offers a convenient expression for homography computation with four point correspondences, its application in homography parameterization and neural network estimation remains unexplored. 

\begin{figure*}[t]
\begin{center}
\includegraphics[width=\textwidth]{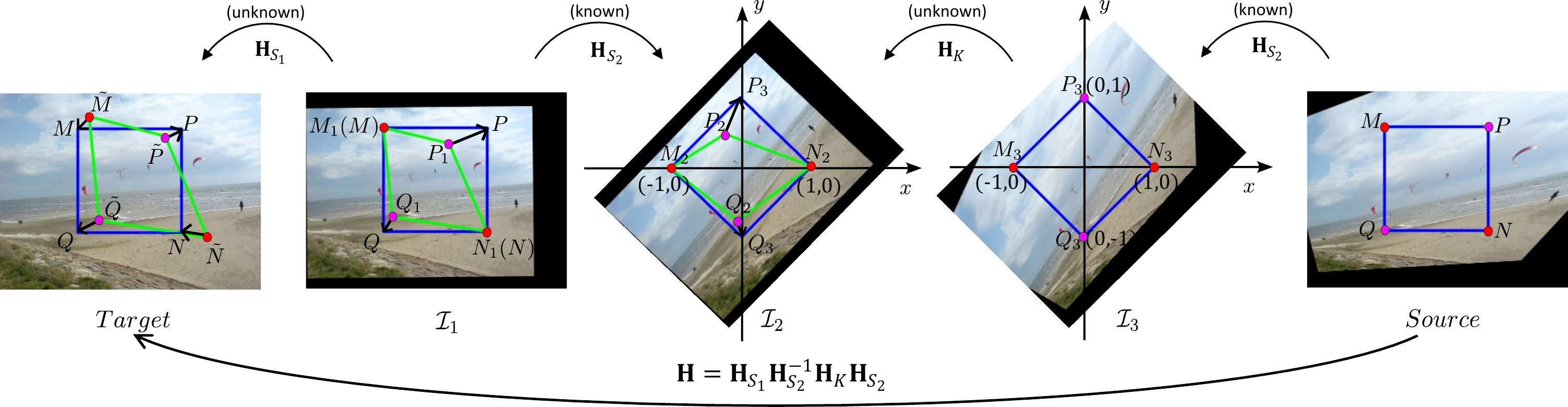}
\end{center}
\vspace{-4mm}
\caption{Improved SKS decomposition. Among the four sub-transformations, the similarity transformation \( \mathbf{H}_{S_2}\) that normalizes $\{M, N\}$ to points $[\mp1,0]^\top$ is known. The unknown similarity transformation \( \mathbf{H}_{S_1}\) plays the role of mapping $\{M, N\}$ to $\{\tilde{M}, \tilde{N}\}$. Additionally, the unknown kernel transformation \( \mathbf{H}_{K}\) introduces the 4-DOF geometric distortion between two normalized images $\mathcal{I}_2$ and $\mathcal{I}_3$.}
\label{fig:homo_decomposition}
\vspace{-1mm}
\end{figure*}

\section{Method}
\vspace{-1mm}

\subsection{Geometric Parameterization of Similarity Transformation}
\label{sec:similarity}
\vspace{-1mm}

Before delving into the geometric parameterization of projective transformations, we theoretically analyze why the geometric parameters estimation for similarity transformations are accurate. Since similarity transformations have four DOF, they can be determined by two pairs of corresponding points between images. Drawing on previous deep estimation works, we design the similarity transformation estimation process depicted in~\cref{fig:similarity_estimation}. Two square sub-images with the same size and positions are randomly sampled in the source and target images, respectively. $\mathbf{H}_T$ denotes a special translation transformation that normalizes two squares by centering their origins at the coordinate system's origin in the intermediate images $\mathcal{I}_1$ and $\mathcal{I}_2$. The unknown similarity transformation $\mathbf{H}_S$ from $\mathcal{I}_2$ to $\mathcal{I}_1$ is to be solved. A deployed neural network takes the two square sub-images as inputs. One choice of the output is the four offsets of two corners (e.g., the red diagonal corners $M$ and $N$ are chosen). Our prediction, however, selects the four unknown parameters in $\mathbf{H}_S$. The entire similarity transformation $\bar{\mathbf{H}}_S$ from the source image to the target image is decomposed by
\begin{equation}
    \bar{\mathbf{H}}_S = \mathbf{H}_{T}^{-1}{\mathbf{H}}_S\mathbf{H}_{T},
\end{equation}
where the unknown ${\mathbf{H}}_S$ is expressed by 
\begin{equation}
\mathbf{H}_S = \begin{bmatrix}
\Delta a_{S}+1 & -b_{S}  & u_{S} \\
b_{S}   & \Delta a_{S}+1 & v_{S} \\
  0     &  0      & 1 
\end{bmatrix}.
\label{equ:HS_expression}
\end{equation}

Taking $\{\!M\!\!\rightarrow\!\!{\tilde{M}}\!\}$ as an example, the offset is defined as
\begin{equation}
\begin{split}
\begin{bmatrix}
\Delta x_{M} \\
\Delta y_{M}  \\
0
\end{bmatrix} = \begin{bmatrix}
x_{M} - x_{\tilde{M}}  \\
y_{M} - y_{\tilde{M}} \\
0
\end{bmatrix} = \begin{bmatrix}
x_{M} \\
y_{M} \\
1
\end{bmatrix} - \bar{\mathbf{H}}_S\begin{bmatrix}
x_{M} \\
y_{M} \\
1
\end{bmatrix},
\end{split}
\end{equation}
where $[x_{M}, y_{M}]^\top$ and $[x_{\tilde{M}}, y_{\tilde{M}}]^\top$ are the image coordinates of the points $M$ and $\tilde{M}$, respectively. Since $\mathbf{H}_T$ and $\mathbf{H}_T^{-1}$ counteract the respective displacements of the corners between the source and target images, we have 
\begin{equation}
\begin{bmatrix}
\Delta x_{M} \\
\Delta y_{M}  \\
0
\end{bmatrix} = \begin{bmatrix}
x_{M_2} - x_{M_1}  \\
y_{M_2} - y_{M_1} \\
0
\end{bmatrix} = \begin{bmatrix}
-r \\
r \\
1
\end{bmatrix} - {\mathbf{H}}_S\begin{bmatrix}
-r \\
r \\
1
\end{bmatrix},
\end{equation}
where $r$ denotes half the length of the square's side.

Subsequently, the offsets of the two corners can be determined by the parameters of the similarity transformation $\mathbf{H}_S$, as shown in the following equation
\vspace{-1mm}
\begin{alignat}{6}
    &\Delta x_{M} \;&=&\; &r*\Delta a_{S} \;&+&\; &r*b_{S}  \;&-\; u_{S},\\
    &\Delta y_{M} \;&=&\; (-r&)*\Delta a_{S} \;&+&\; &r*b_{S}  \;&-\; v_{S},\\
    &\Delta x_{N} \;&=&\; (-r&)*\Delta a_{S} \;&+&\; (-r&)*b_{S}  \;&-\; u_{S},\\
    &\Delta y_{N} \;&=&\; &r*\Delta a_{S} \;&+&\; (-r&)*b_{S}  \;&-\; v_{S},
    \vspace{-4mm}
\end{alignat}
where if $\bar{\mathbf{H}}_S$ represents an identity transformation, then all four positional offsets will be zero, as will the four geometric parameters that we adopt.

It is evident that these two kinds of parameterization for the similarity transformation are equivalent through a simple linear transformation. Such linear transformations can be effectively modeled using a single-layer multi-layer perceptron (MLP) network. Consequently, the estimation results from these two parameterizations should be nearly identical, provided that the mapping ranges are appropriately defined or normalized. Compared to the parameterization with four positional offsets, which requires solving a linear system to express $\mathbf{H}_S$, the geometric parameterization is more straightforward.

Once \(\mathbf{H}_S\) is estimated from the deployed neural network, the whole similarity transformation \(\bar{\mathbf{H}}_S\) can be obtained through matrix multiplication. The scale and rotation remain unchanged, i.e., \(\Delta \bar{a}_S = \Delta a_S\) and \(\bar{b}_S = b_S\). The reason we do not directly estimate the geometric parameters of \(\bar{\mathbf{H}}_S\) is that its translation components (denoted by \(\bar{u}_S\) and \(\bar{v}_S\)) are affected by disturbances in the position and side length of the square, as well as by \(\Delta a_S\) and \(b_S\). Consequently, the conversion between positional offsets and geometric parameters of \(\bar{\mathbf{H}}_S\) becomes more complex.

For affine transformations, the estimation process is similar to that illustrated in~\cref{fig:similarity_estimation}. The difference lies in the selection of three square corners, and the intermediate similarity transformation is replaced by an affine transformation. As a result, the positional offsets of the three corners and the six geometric parameters of the affine transformation remain equivalent, connected through a linear transformation, similar to Eqs.~\textcolor{iccvblue}{7} to \textcolor{iccvblue}{10}.

\subsection{Improved SKS for Homography Parameterization}
\label{sec:solving}
\vspace{-1mm}

The original SKS approach outlined in~\cref{subsec:SKS} decomposes a homography into three sub-transformations. To effectively apply the SKS decomposition to deep homography estimation, we propose an improved SKS that ensures each sub-transformation is suitable for neural network fitting, as depicted in~\cref{fig:homo_decomposition}. 

Within this framework, \( \mathbf{H}_{S_2}\) is pre-computed, imposing the translation, scaling, and rotation to normalize $\{M, N\}$ to the canonical coordinates $[\mp1,0]^\top$. Under \( \mathbf{H}_{S_2}\), the selected square in $\mathcal{I}_1$ and the source image are transformed to a diamond shape~\cite{wolfram_diamond} in $\mathcal{I}_2$ and $\mathcal{I}_3$, respectively. The role of \( \mathbf{H}_{S_1}\) corresponds to the entire similarity transformation \(\bar{\mathbf{H}}_S\) depicted in~\cref{fig:similarity_estimation}, mapping the corners $\{M, N\}$ in the source image to $\{\tilde{M}, \tilde{N}\}$ in the target image. The four geometric parameters associated with \( \mathbf{H}_{S_1}\) are still represented as the four parameters in the middle similarity transformation \( \mathbf{H}_{S}\). Therefore, 
the complete homography from the source image to the target image is represented by 
\begin{equation}
\resizebox{0.88\linewidth}{!}{%
$
    {\mathbf{H}} =  {\mathbf{H}}_{S_1} {\mathbf{H}}_{S_2}^{-1} {\mathbf{H}}_{K} {\mathbf{H}}_{S_2} = \mathbf{H}_{T}^{-1} {\mathbf{H}}_S  \mathbf{H}_{T} {\mathbf{H}}_{S_2}^{-1} {\mathbf{H}}_{K} {\mathbf{H}}_{S_2},
    $}
\end{equation}
where ${\mathbf{H}}_S$ is given as~\cref{equ:HS_expression} and ${\mathbf{H}}_K$ is expressed by
\begin{equation}
\mathbf{H}_K = \begin{bmatrix}
\Delta a_{K}+1 & u_{K}  & b_{K} \\
  0     &  1     & 0 \\
b_{K}   & v_{K} & \Delta a_{K}+1
\end{bmatrix}.
\label{equ:HK_expression}
\end{equation}

We now introduce a 8-DOF geometric parameterization for homography, which is decoupled into two independent sets: 4-DOF in ${\mathbf{H}}_S$ and 4-DOF in ${\mathbf{H}}_K$. The four geometric parameters in ${\mathbf{H}}_S$ have been shown to be linearly related to the four positional offsets of $M$ and $N$, as indicated in Eqs.~\textcolor{iccvblue}{7} to \textcolor{iccvblue}{10}. Consequently, the four offsets of the other two corners can be expressed as follows.

Take the offsets from $\tilde{P}$ to ${P}$ as an example. 
The point $\tilde{P}$ is expressed as a series of multiplications 
\begin{equation}
\begin{split}
\begin{bmatrix}
x_{\tilde{P}} \\
y_{\tilde{P}} \\
1
\end{bmatrix} &= \mathbf{H}_{S_1}\begin{bmatrix}
x_{P_1} \\
y_{P_1} \\
1
\end{bmatrix} = \mathbf{H}_{S_1}\mathbf{H}_{S_2}^{-1}\begin{bmatrix}
x_{P_2} \\
y_{P_2} \\
1
\end{bmatrix} \\
&= \mathbf{H}_{T}^{-1} {\mathbf{H}}_S \mathbf{H}_{T} \mathbf{H}_{S_2}^{-1}\mathbf{H}_{K}\begin{bmatrix}
x_{P_3} \\
y_{P_3} \\
1
\end{bmatrix}.
\end{split}
\vspace{-1mm}
\end{equation}

By substituting all known variables into these expressions and performing some simplifications, the involved two offsets are represented by  
\begin{equation}
\small
\begin{split}
 \Delta x_{P} = &r*\frac{(\Delta a_S + 1 + b_S)*(b_K + u_K) + (\Delta a_S + 1 - b_S)}{\Delta a_k + 1 + v_K} \\ &- \Delta a_S*x_O + b_S*y_O + u_S -r\,,  \\
    \Delta y_{P} = &r*\frac{(b_S - \Delta a_S - 1)*(b_K + u_K) + (\Delta a_S + 1 + b_S)}{\Delta a_k + 1 + v_K} \\ &- \Delta a_S*y_O- b_S*x_O + v_S - r\,. 
\end{split}
\end{equation}

\vspace{-2mm}
Similarly, $\tilde{Q}$ can also be expressed in the same manner. It is observed that these offsets are complex, incorporating the fractional representation of eight geometric parameters. While this expression offers insights into the ranges of values and simple divisions within the parameter space, it remains challenging to theoretically prove that neural networks can estimate these parameters. The main problem lies in whether ${\mathbf{H}}_K$ can be effectively estimated. In the following section, we will explore the geometric interpretation of ${\mathbf{H}}_K$ and demonstrate why its parameters can be reliably estimated from the two images.

\subsection{Geometric Interpretation of Kernel Transformation}
\label{subsec:Geo_mean_HK}
\vspace{-1mm}

The geometric interpretation of the four parameters in the kernel transformation ${\mathbf{H}}_K$ has not been explored in the original SKS research~\cite{Mine_SKS-I}. In alignment with the four geometric parameters of ${\mathbf{H}}_S$ discussed in~\cref{sec:similarity}, this subsection aims to demonstrate the geometric significance of $\{a_K, b_K, u_K, v_K\}$ in the context of projective distortion. As illustrated in~\cref{fig:homo_decomposition}, under the kernel transformation, the canonical points \(P_3(0,1)\) and \(Q_3(0,-1)\) in $\mathcal{I}_3$ are mapped to the points \(P_2\) and \(Q_2\) in $\mathcal{I}_2$, respectively. Specifically, the coordinates of $P_2$ and $Q_2$ are expressed as
\begin{equation}
\begin{bmatrix}
x_{P_2} \\
y_{P_2}
\end{bmatrix} \!=\! \begin{bmatrix}
\frac{b_K+u_K}{\Delta a_K + v_K + 1} \\
\frac{1}{\Delta a_K + v_K + 1} 
\end{bmatrix}, \; \begin{bmatrix}
x_{Q_2} \\
y_{Q_2}
\end{bmatrix} \!=\! \begin{bmatrix}
\frac{b_K-u_K}{\Delta a_K - v_K + 1} \\
\frac{-1}{\Delta a_K - v_K + 1}
\end{bmatrix}.
\end{equation}

\begin{figure}[t]
\begin{center}
\includegraphics[width=0.96\linewidth]{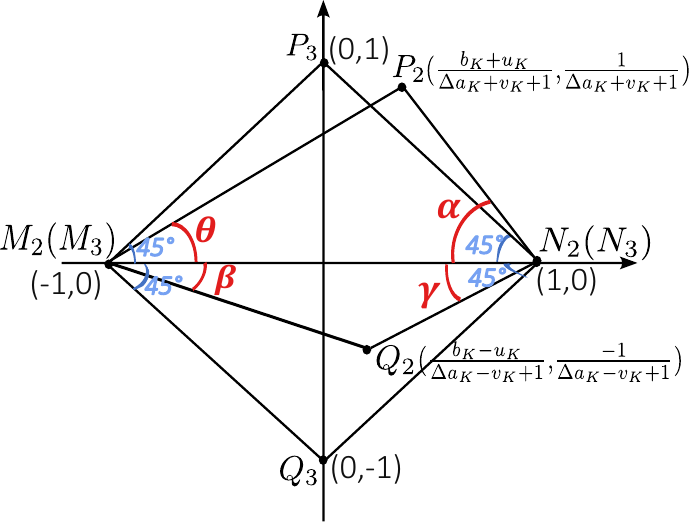}
\end{center}
\vspace{-5mm}
\caption{Our geometric interpretation of kernel transformation. The proposed four angular offsets explicitly define the 4-DOF geometric distortion induced by the kernel transformation.}
\label{fig:Geo_mean_HK}
\vspace{-2mm}
\end{figure}

This geometric distortion induced by ${\mathbf{H}}_K$ is further illustrated in~\cref{fig:Geo_mean_HK}. Through careful observation and derivation, we establish a linear transformation that relates the cotangent values of the four angles depicted in the figure to the four geometric parameters in ${\mathbf{H}}_K$. 

For example, consider the angle $\theta$ between $M_2P_2$ and $M_2N_2$. The corresponding angle between $M_3P_3$ and $M_3N_3$ is $45^{\circ}$. The cotangent of $\theta$ can be expressed in terms of the coordinates of $M_2$ and $P_2$:
\begin{equation}
\begin{split}
\cot\theta &= \frac{x_{P_2} - x_{M_2}}{y_{P_2}} = \Delta a_K + b_K + u_K + v_K + 1.
\end{split}
\end{equation}

Consequently, we define the following angular offsets:
\vspace{-1mm}
\begin{small}
\begin{alignat}{4}
&\Delta \cot \theta \;&=&\; \cot \theta - \cot 45^{\circ} \;&=&\; \Delta a_K + b_K + u_K + v_K,\\
&\Delta \cot \alpha \;&=&\;  \cot \alpha - \cot 45^{\circ} \;&=&\; \Delta a_K - b_K - u_K + v_K, \\
&\Delta \cot \beta \;&=&\;  \cot \beta - \cot 45^{\circ} \;&=&\; \Delta a_K + b_K - u_K - v_K, \\
&\Delta \cot \gamma \;&=&\;  \cot \gamma - \cot 45^{\circ} \;&=&\; \Delta a_K - b_K + u_K - v_K,
\vspace{-1mm}
\end{alignat}
\end{small}

\noindent where if ${\mathbf{H}}_K$ denotes an identity transformation, then all four angular offsets will be zero, as will the four geometric parameters of the kernel transformation we propose.

\begin{figure}[t]
\begin{center}
\includegraphics[width=1.0\linewidth]{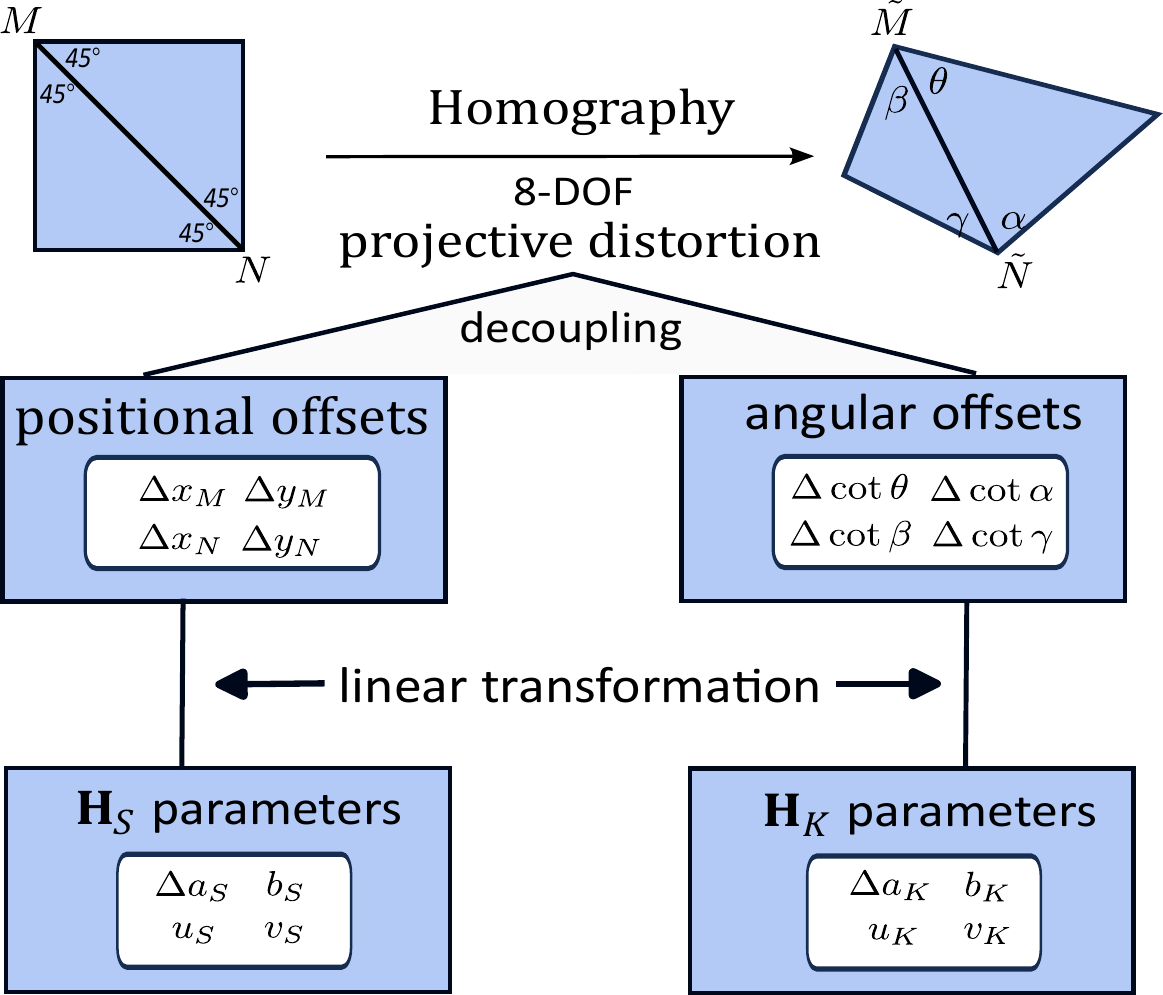}
\end{center}
\vspace{-4mm}
\caption{Decoupled projective distortion and geometric parameters of homography. Under homography, a quadrangle (here is a square) undergoes an 8-DOF projective distortion, which we have decoupled into four positional and four angular offsets. The two sets of four geometric parameters we propose are demonstrated to be linearly transformable into their respective sets of offsets.}
\label{fig:8DOF_summary}
\vspace{-3mm}
\end{figure}

It is worth noting that the defined four angular offsets remain unchanged under similarity transformations. In other words, they offer a novel metric for evaluating the entire projective distortion between the source square (in blue) and the target quadrangle (in green) shown in~\cref{fig:homo_decomposition}, independent of the first four geometric parameters in \(\mathbf{H}_{S_1}\). Consequently, we have demonstrated that the two sets of decoupled four parameters in SKS correspond to two distinct sets of decoupled projective distortion meanings. As a result, the angular offsets we propose can be effectively estimated by neural networks, akin to the positional offsets.~\Cref{fig:8DOF_summary} sketches the proposed parameterization for an 8-DOF homography.

\subsection{Application in Deep Homography Estimation}
\label{subsec:loss}
\vspace{-1mm}

One application of our 8-DOF geometric parameterization is its seamless integration into existing neural networks for homography estimation, where it replaces the commonly used four-corner positional offsets as the output. For non-iterative homography estimation methods, such as~\cite{DHN16,RAL18,UDHN}, the training procedure requires only the modification in the loss function. For iterative homography estimation methods, such as those discussed in~\cite{MHN,Nowruzi2017,IHN,MCNet}, the homography computation will shift from solving a linear system based on four-corner positional offsets to performing matrix multiplication. This change not only simplifies the implementation but also enhances computational efficiency.

More importantly, the proposed geometric parameterization provides a unified solution for 2D transformation estimation. Specifically, by analyzing the four angles \{$\theta$, $\alpha$, $\beta$, $\gamma$\}, one can directly identify the parallelogram in affine distortion, as $\theta$$=$$\gamma$ and $\alpha$$=$$\beta$. Under a 4-DOF similarity transformation, the four angles \{$\theta$, $\alpha$, $\beta$, $\gamma$\} in Fig.~\ref{fig:8DOF_summary} will remain unchanged at $45^\circ$. In terms of 8-DOF geometric parameters, the following criterion is established by us to identify the degenerate affine and similarity transformations:

\vspace{1mm}
\noindent \textbf{Criterion of identifying affine and similarity transformations in geometric parameters}. \textit{A homography with the proposed geometric parameterization degenerates to an affine transformation, if and only if both $b_K$ and $v_K$ are zero. Furthermore, the homography degenerates to a similarity transformation, if and only if $b_K$, $v_K$, $\Delta a_K$, and $u_K$, are all zero.}

In contrast, the four-point positional offset parameterization requires solving for the homography matrix first, then decomposing it (e.g., using the methods in~\cite{Hartley2003Multiple} [p.~42–43]) to extract the projective and affine components, which can be cumbersome.

\section{Experiments}
\vspace{-1mm}

\subsection{Configurations}
\label{sec:conf}
\vspace{-1mm}

\textbf{Datasets:} We evaluate the homography estimation task across three commonly used datasets, including one dynamic scene dataset and one cross-modal dataset. A brief description of each dataset is provided below:
\begin{itemize}
    \item \textbf{MSCOCO~\cite{MSCOCO}:} A large-scale image dataset widely used for evaluating homography estimation methods, consisted of synthetic image pairs.
    \item \textbf{SPID~\cite{SPID}:} A dataset captured in real-world surveillance scenes, consisted of real image pairs which include pedestrian occlusion and varying lighting conditions. 
    \item \textbf{GoogleMaps~\cite{CLKN}:} A real dataset consisted of a training set of 8,822 paired static Google Maps images and satellite images, along with a test set of 888 paired images.
    
\end{itemize}
\vspace{1mm}
\textbf{Networks:} We conduct tests across multiple homography estimation networks that predict eight parameters simultaneously, as listed below. Traditional methods are also included for comparison.
\begin{itemize}
    \item \textbf{DHN~\cite{DHN16}:} DHN is a pioneering deep learning-based homography estimation method that uses a convolutional neural network to predict homographies.
    \item \textbf{RHWF~\cite{RHWF}:} Recurrent homography estimation utilizes homography-guided image warping and a focus transformer to iteratively refine the estimation results.
    \item \textbf{MCNet~\cite{MCNet}:} MCNet is the recent SOTA approach that enhances homography estimation accuracy and efficiency through multi-scale correlation searching.
    \item \textbf{SIFT~\cite{SIFT} + RANSAC~\cite{RANSAC} / MAGSAC~\cite{MAGSAC}}: SIFT is a classical keypoint extraction and matching algorithm; RANSAC/MAGSAC are hypothesis-verification techniques used to estimate models with outliers.
\end{itemize}
\vspace{1mm}

\begin{figure*}[htbp]
    \centering
    \begin{subfigure}{0.32\textwidth}
        \centering
        \includegraphics[width=\textwidth]{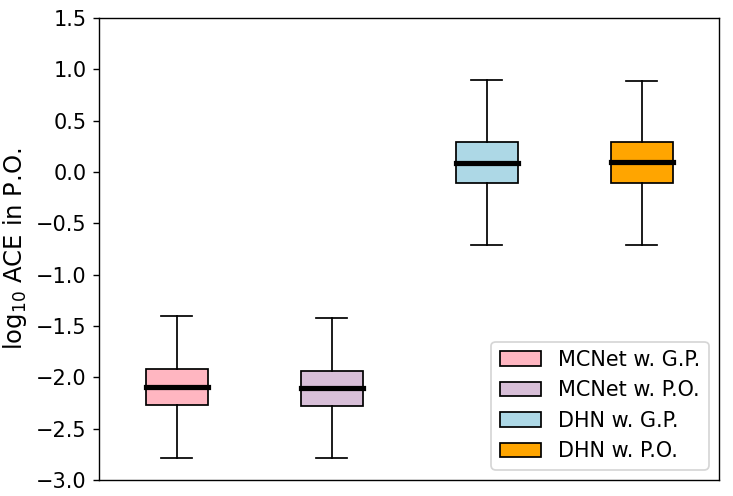}
         \vspace{-5mm}
        \caption{MSCOCO}
    \end{subfigure} \,
    \begin{subfigure}{0.32\textwidth}
        \centering
        \includegraphics[width=\textwidth]{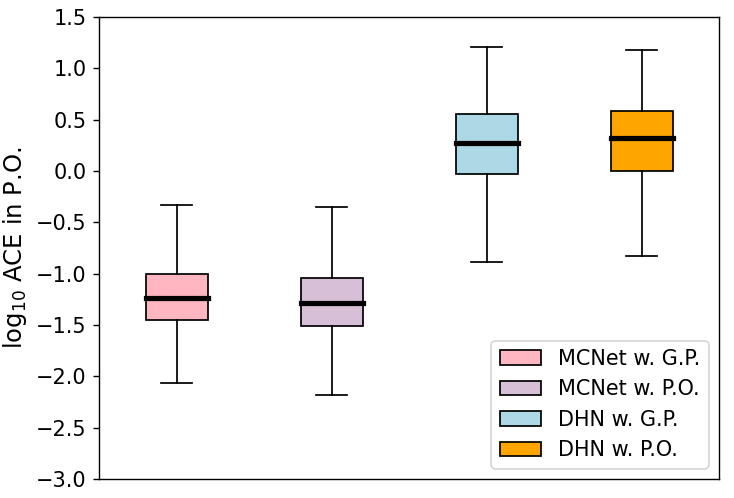}
         \vspace{-5mm}
        \caption{SPID}
    \end{subfigure} \,
    \begin{subfigure}{0.32\textwidth}
        \centering
        \includegraphics[width=\textwidth]{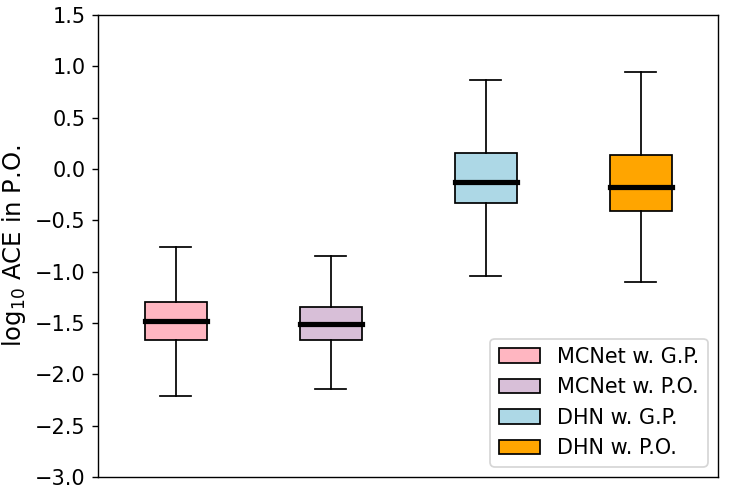}
         \vspace{-5mm}
        \caption{GoogleMap}
    \end{subfigure}
    \vspace{-3mm}
    \caption{Quartile plots of ACE in P.O. for similarity estimation. The four geometric parameters (G.P.) and the two-corner positional offset (P.O.) parameterization are evaluated on three distinct datasets, using DHN~\cite{DHN16} and MCNet~\cite{MCNet}.}
    \label{fig:similarityError}
    \vspace{-2mm}
\end{figure*}



\begin{figure*}[htbp]
    \centering
        \begin{subfigure}{0.32\textwidth}
            \centering
            \includegraphics[width=\textwidth]{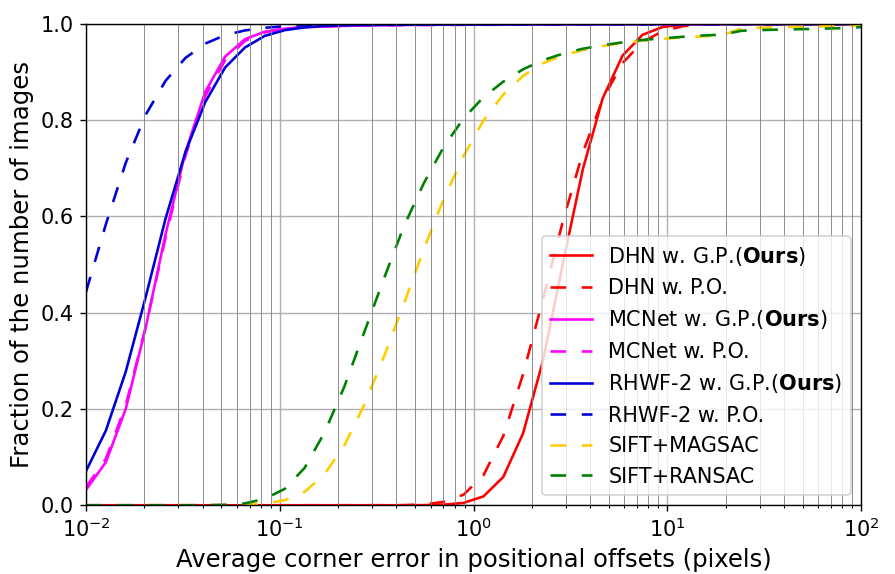}
            \caption{MSCOCO}
        \end{subfigure} \,
        \begin{subfigure}{0.32\textwidth}
            \centering
            \includegraphics[width=\textwidth]{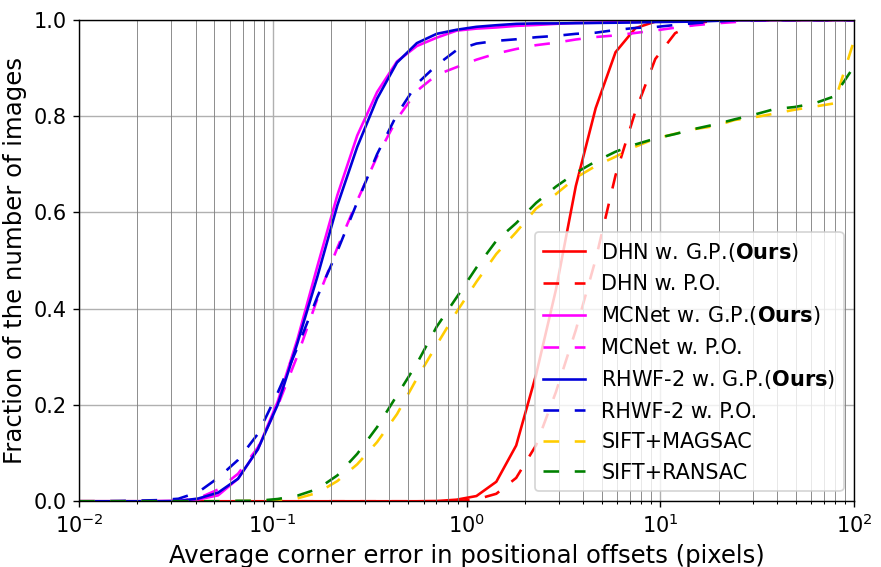}
            \caption{SPID}
        \end{subfigure} \,
        \begin{subfigure}{0.32\textwidth}
            \centering
            \includegraphics[width=\textwidth]{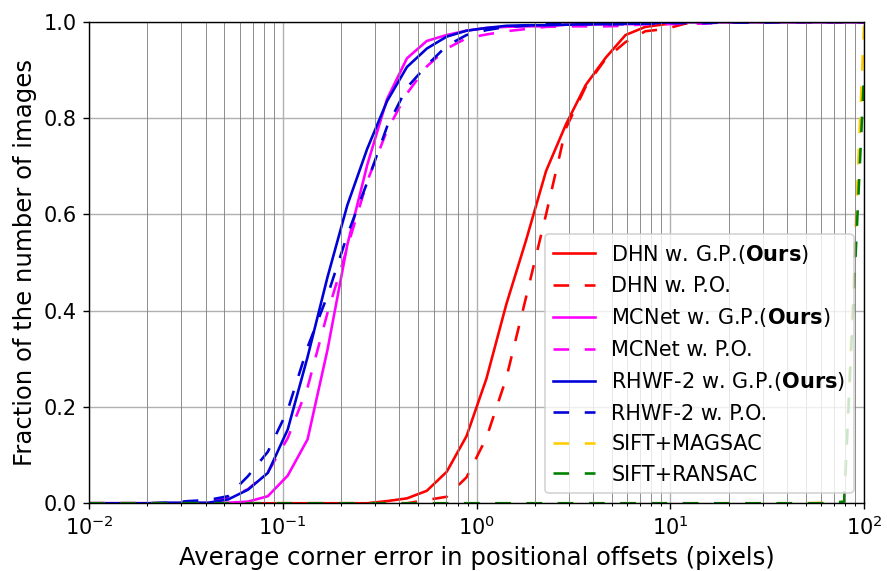}
            \caption{GoogleMap}
        \end{subfigure}
    \vspace{-3mm}
    \caption{Homography estimation results on MSCOCO, dynamic SPID, and cross-modal GoogleMap datasets, with the proposed geometric parameters (G.P.) and the four-corner positional offsets (P.O.) parameterization for deep learning methods.}
    \label{fig:Homo_comp}
    \vspace{-3mm}
\end{figure*}

\noindent \textbf{Data Generation:} For the datasets mentioned above, we apply the same image processing technique used in previous methods~\cite{DHN16,Nowruzi2017,MHN,LocalTrans,CLKN,DLKN, IHN,RHWF,MCNet} to generate data for homography estimation. Specifically, random perturbations are introduced at the four corner points of the input $128\!*\!128$ images, with perturbations ranging from -32 to 32 pixels. This approach enables the generation of image pairs with varying degrees of projective distortion. For verifying similarity transformations, we perturb two corners along the main diagonal to generate the image pairs.

\vspace{1mm}
\noindent \textbf{Metric:} To measure the accuracy of deep transformation estimation, we employ average corner error in positional offsets (ACE in P.O.) as the primary evaluation metric. This metric provides an intuitive measure of how accurately the projective distortion between image pairs is reconstructed. Additionally, the average corner error in the four angular offsets (ACE in A.O.), as illustrated in~\cref{fig:8DOF_summary}, is used to assess the accuracy from another perspective.

\textbf{The results presented in this work are fully reproducible using the provided source codes in the Supplementary Material.}

\subsection{Evaluation for Similarity Transformation}
\label{sec:sim_res}
\vspace{-1mm}

We first test the two types of parameterizations of similarity transformation, as described in~\cref{sec:similarity}: two-corner positional offsets and four geometric parameters. The evaluation is conducted on the aforementioned datasets. Quartile plots of the ACE in P.O. for both parameterizations of similarity transformation are depicted in~\cref{fig:similarityError}. Since the image pairs in MSCOCO are generated through an ideal homography, the ACE in P.O. on this dataset is significantly lower than them in the other two datasets. Across three distinct datasets and two neural networks, the results demonstrate that the four geometric parameters perform on par with the two-corner positional offsets, with no noticeable difference.
\begin{table}[t]
\centering
\resizebox{0.48\textwidth}{!}{%
\begin{tabular}{|c|c|c|c|}
\hline
\multirow{2}{*}{Network} & \multicolumn{3}{c|}{{Parameterization: \; P. O. $\longrightarrow$  G.P. (\textbf{Ours})}}  \\ \cline{2-4} 
  &{MSCOCO} & {SPID} & {GoogleMap} \\ \hline
DHN~\cite{DHN16}          & 0.0607 $\rightarrow$ 0.0594 (\textcolor{red}{-2\%})\;  & 0.0680 $\rightarrow$ 0.0547 (\textcolor{red}{-20\%}) & 0.0252 $\rightarrow$ 0.0208 (\textcolor{red}{-18\%})  \\ \hline
RHWF-2~\cite{RHWF}        & 0.0003 $\rightarrow$ 0.0003 (\textcolor{red}{-6\%})\;  & 0.0055 $\rightarrow$ 0.0024 (\textcolor{red}{-56\%}) & 0.0063 $\rightarrow$ 0.0026 (\textcolor{red}{-59\%})      \\ \hline
MCNet~\cite{MCNet}        & 0.0008 $\rightarrow$ 0.0003 (\textcolor{red}{-63\%}) & 0.0058 $\rightarrow$ 0.0022 (\textcolor{red}{-62\%}) & 0.0065 $\rightarrow$ 0.0031 (\textcolor{red}{-52\%}) \\ \hline
\end{tabular}%
}
\vspace{-2mm}
\caption{Switch in homography parameterization from the four-corner positional offsets (P.O.) and our geometric parameters (G.P.) on the median error of four angular offsets. The changes in errors are highlighted in \textcolor{red}{red} to indicate a decrease.}
\label{tab:angularOffsets}
\vspace{-3mm}
\end{table}

\begin{figure*}[h]
\begin{center}
\includegraphics[width=1.0\textwidth]{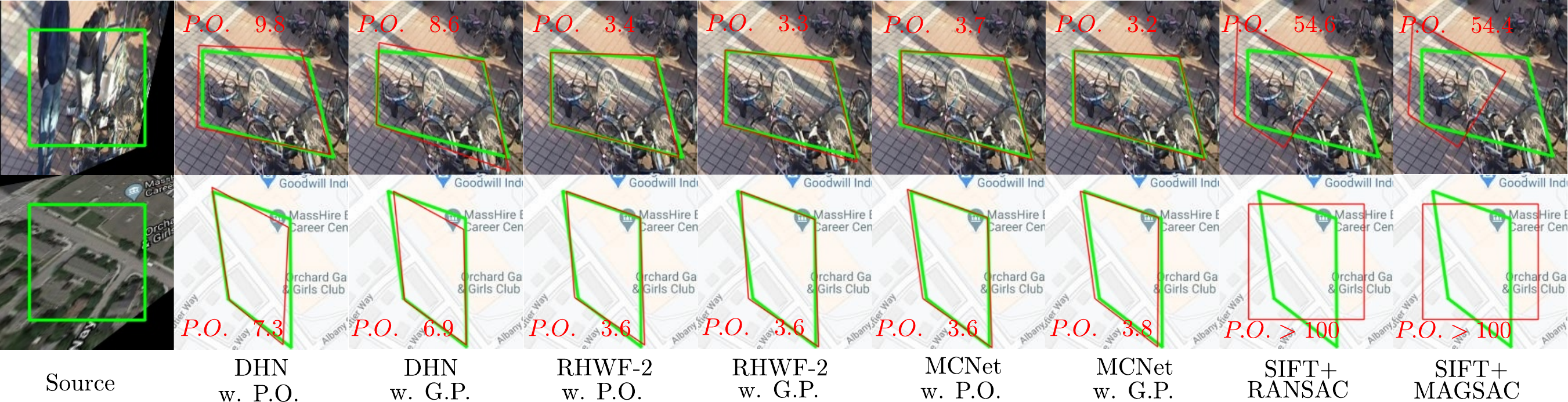}
\end{center}
\vspace{-5mm}
\caption{Challenging cases of homography estimation in the SPID (\textbf{above}) and GoogleMap (\textbf{below}) datasets. The green and red quadrangles represent the ground-truth locations and predicted locations estimated by different algorithms, respectively. The errors, marked in red in each sub-figure, are evaluated using the positional offsets (P.O.) of four corners.}
\label{fig:imageExample}
\vspace{-1mm}
\end{figure*}
\begin{figure*}[t]
    \centering
    \begin{subfigure}{0.32\textwidth}
        \centering
        \includegraphics[width=\textwidth]{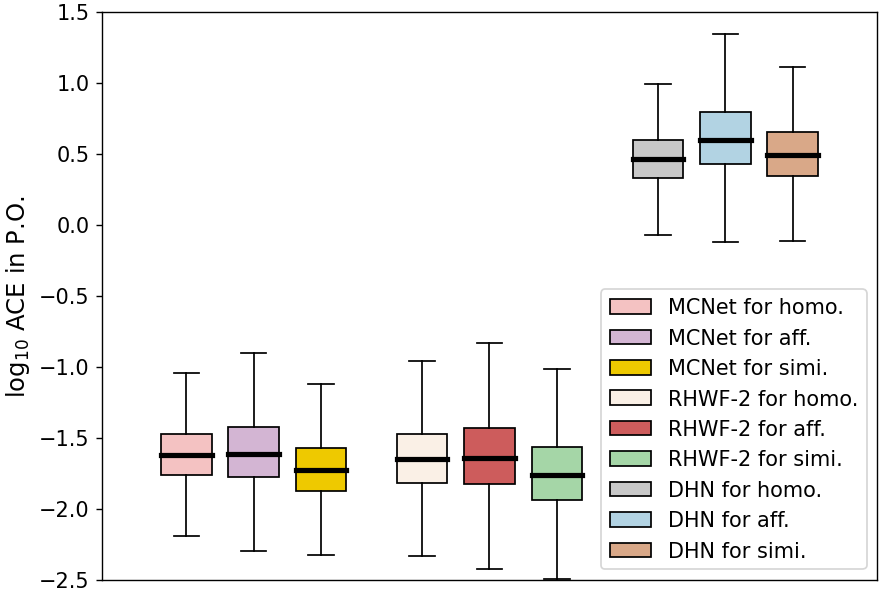}
         \vspace{-4mm}
        \caption{MSCOCO}
    \end{subfigure} \,
    \begin{subfigure}{0.32\textwidth}
        \centering
        \includegraphics[width=\textwidth]{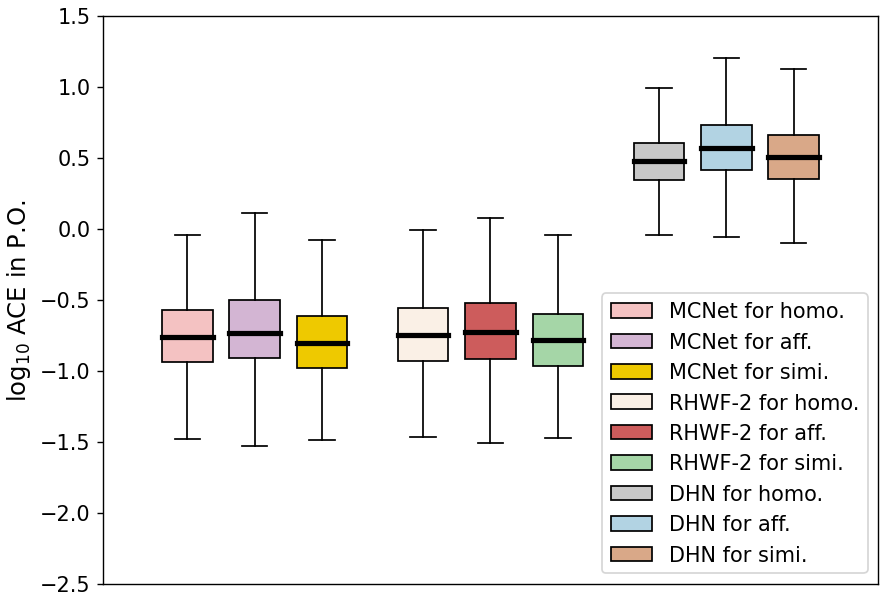}
         \vspace{-4mm}
        \caption{SPID}
    \end{subfigure} \,
    \begin{subfigure}{0.32\textwidth}
        \centering
        \includegraphics[width=\textwidth]{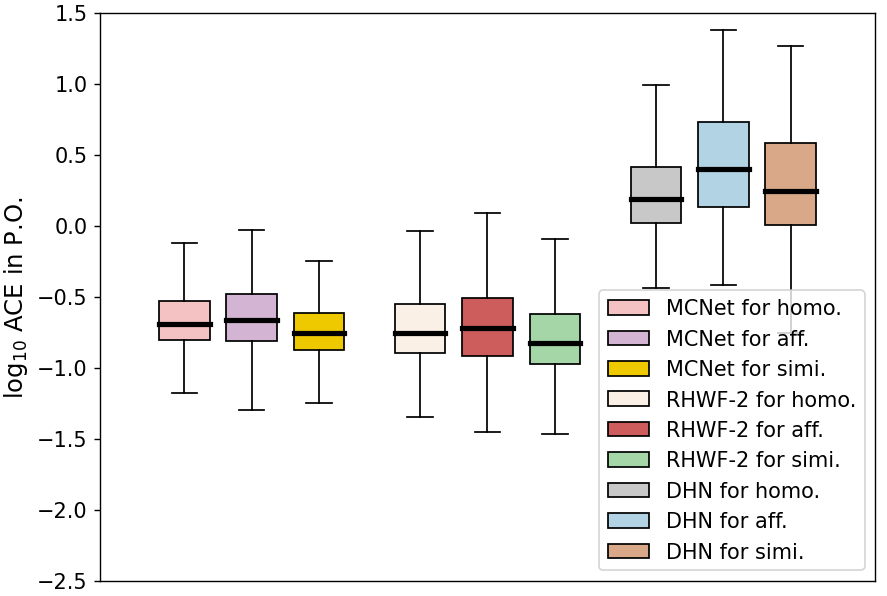}
         \vspace{-4mm}
        \caption{GoogleMap}
    \end{subfigure}
    \vspace{-3mm}
    \caption{Quartile plots of ACE in P.O. for simultaneous estimation of homography, affine, and similarity using eight geometric parameters. The evaluations are conducted on commonly used MSCOCO, dynamic SPID, and cross-modal GoogleMap datasets, generated under homography, affine, similarity, respectively. Three distinct networks are used for integration: DHN~\cite{DHN16}, RHWF-2~\cite{RHWF}, and MCNet~\cite{MCNet}.}
    \label{fig:ACEresults}
    \vspace{-1mm}
\end{figure*}

\vspace{-4mm}
\subsection{Evaluation for Homography}
\label{sec:exp_runtime}
\vspace{-1mm}

The two homography parameterizations are evaluated across four distinct datasets and three network architectures. As depicted in~\cref{fig:Homo_comp}, the overall results demonstrate that the proposed geometric parameterization (G.P.) performs on par with the common four-corner positional offsets (P.O.), with only slight differences across datasets.

\begin{table}[t]
\centering
\resizebox{0.48\textwidth}{!}{%
\begin{tabular}{|c|c|c|c|}
\hline
    {Ablation} & {Setting} & \tabincell{c}{ Parameterization: \\ P. O. $\rightarrow$ G.P. (\textbf{Ours})} & \tabincell{c}{{Inference} \\ {Time (ms)}} \\ \hline
\multirow{3}{*}{Iteration}  & (2, 2, 2) & 0.0240 $\rightarrow$ 0.0239 (\textcolor{red}{-0.4\%})\;  & 21.0 \\ \cline{2-4}
                            & (4, 4, 4) & 0.0228 $\rightarrow$ 0.0223 (\textcolor{red}{-2.2\%})  & 35.1 \\ \cline{2-4}
                            & (8, 8, 8) & 0.0197 $\rightarrow$ 0.0180 (\textcolor{red}{-8.6\%})  & 62.4 \\ \hline
\multirow{3}{*}{Scale}      & 1         & 0.3312 $\rightarrow$ 0.3245 (\textcolor{red}{-2.0\%})  & 20.3 \\ \cline{2-4}
                            & 2         & 0.0531 $\rightarrow$ 0.0486 (\textcolor{red}{-8.5\%})  & 20.5 \\ \cline{2-4}
                            & 3         & 0.0240 $\rightarrow$ 0.0239 (\textcolor{red}{$-0.4\%$})  & 21.0 \\ \hline
\end{tabular}
}
\vspace{-2mm}
\caption{Ablation on our geometric parameterization (G.P.) and four-corner positional offset (P.O.) parameterization under different configurations of MCNet~\cite{MCNet}. The percentage decrease in median of ACE in P.O. is highlighted in \textcolor{red}{red}.}
\label{tab:ablation}
\vspace{-3mm}
\end{table}

\Cref{fig:imageExample} illustrates two challenging homography estimation cases. The recent MCNet and RHWF with G.P. perform similarly to them with P.O., but outperform other methods.~\Cref{tab:angularOffsets} further provides the median of average corner error in the four angular offsets (ACE in A.O.), offering an alternative perspective on the accuracy of homography estimation. The results show notable differences compared to the ACE in P.O. shown in~\cref{fig:Homo_comp}. Among the nine combinations of network architecture and dataset, switching homography parameterization from P.O. to G.P. improves performance across all cases.

\vspace{-1mm}
\subsection{Ablation Study for MCNet's Configurations}
\label{sec:exp_runtime}
\vspace{-1mm}

The recent MCNet~\cite{MCNet} employs multiscale correlation searching in an iterative manner, with default iteration times of (2, 2, 2) across three scales. To thoroughly evaluate the proposed geometric parameterization, we replicate MCNet's ablation configurations. From Table~\ref{tab:ablation}, it is observed that switching the homography parameterization from P.O. to G.P. leads to a slight performance improvement across all MCNet's configurations.

\subsection{Similarity and Affine Estimation with Trained Homography Networks}
\vspace{-1mm}
Since similarity and affine are degenerate cases of projective transformations, we propose transferring the trained neural networks for homography estimation to directly predict the geometric parameters of similarity and affine transformations. Specifically, we use the deep homography networks to predict the eight geometric parameters of affine and similarity transformations, rather than reducing them to their simpler forms. Evaluations are conducted on the aforementioned datasets with the following metric.

\vspace{1mm}
\noindent \textbf{Evaluation with average corner error in positional offsets (ACE in P.O.):} This metric measures how accurately the projective distortion between image pairs is reconstructed under degenerate cases. \Cref{fig:ACEresults} shows quartile plots of the ACE in P.O. for the same parameterization of 2D transformations, including homography, affine, and similarity, respectively. Since DHN’s network architecture is less advanced and processes grayscale image pairs, its ACE in P.O. results are significantly lower than those of the two more recent SOTA methods. Across all three datasets and three neural networks, the results demonstrate that the eight geometric parameters perform well for affine and similarity transformations.

\vspace{-1mm}
\section{Conclusion and Future Work}
\vspace{-1mm}
This paper presents a novel geometric parameterization of homography that is suitable for estimation through neural networks, based on the SKS decomposition. By introducing two independent sets of four geometric parameters, each with corresponding projective distortion interpretations, our parameterization not only aligns with but also unifies the estimation for 2D similarity and affine transformations. Furthermore, the proposed method eliminates the need for solving linear systems, as required by traditional four-corner positional offsets parameterization, and achieves competitive performance across multiple datasets and neural network architectures. Similar to pose estimation and other geometric vision tasks, our approach demonstrates the value of geometric parameterization, as all deep learning methods predicting positional offsets are not end-to-end and require an algebraic solver (which may itself be complex) to compute solutions as a post-processing step.

Moreover, to the best of our knowledge, this is the first work to introduce angular offsets in vision tasks. In the future, we plan to explore other applications of angular offsets, such as keypoint extraction and matching. We also aim to extend the affine-core-affine (ACA) homography decomposition method to multi-plane homography estimation.

\bibliographystyle{ieee}
\bibliography{main_arxiv}



\clearpage

\appendices

\twocolumn[
\begin{@twocolumnfalse}
\section*{\centering{Supplementary Material for \\[5pt] \emph{Decoupled Geometric Parameterization and its Application in Deep Homography Estimation}\\[25pt]}}
\end{@twocolumnfalse}
]

\renewcommand{\theequation}{\thesection.\arabic{equation}}

\renewcommand{\thefigure}{\thesection.\arabic{figure}}
\setcounter{figure}{0}

\renewcommand{\thetable}{\thesection.\arabic{table}}
\setcounter{table}{0}

\renewcommand\thesection{\Alph{section}}

\section{Geometric Parameterization of Affine Transformation}
\label{sec:affine}

\subsection{Geometric Parameterization Equivalent to Positional Offsets of Three Corners}

Similar to the derivation of the similarity transformation outlined in Sec.~\textcolor{red}{3.1} of the main manuscript, the entire affine transformation, $\bar{\mathbf{H}}_A$, from the source image to the target image can be decomposed as follows:
\begin{equation}
    \bar{\mathbf{H}}_A = \mathbf{H}_{T}^{-1}{\mathbf{H}}_A\mathbf{H}_{T},
    \label{equ:H_A_bar}
\end{equation}
where the unknown ${\mathbf{H}}_A$ is expressed by 
\begin{equation}
\mathbf{H}_A = \begin{bmatrix}
\Delta a_{A}+1 & b_{A}  & u_{A} \\
c_{A}   & \Delta d_{A}+1 & v_{A} \\
  0     &  0      & 1 
\end{bmatrix}.
\label{equ:HA_expression}
\end{equation}

Utilizing three point correspondences: $\{M{\stackrel{\bar{\mathbf{H}}_A}{\longrightarrow}}\tilde{M}\}$, \\$\{N{\stackrel{\bar{\mathbf{H}}_A}{\longrightarrow}}\tilde{N}\}$, and $\{P{\stackrel{\bar{\mathbf{H}}_A}{\longrightarrow}}\tilde{P}\}$, the following expression is obtained:
\begin{equation}
\resizebox{0.86\linewidth}{!}{$
\begin{bmatrix}
\Delta x_{M} \\
\Delta y_{M} \\
\Delta x_{N} \\
\Delta y_{N} \\
\Delta x_{P} \\
\Delta y_{P} \\
\end{bmatrix}
=
\begin{bmatrix}
r & -r & 0 & 0 & -1 & 0 \\
0 & 0 & r & -r & 0 & -1 \\
-r & r & 0 & 0 & 1 & 0 \\
0 & 0 & -r & r & 0 & 1 \\
-r & -r & 0 & 0 & -1 & 0 \\
0 & 0 & -r & -r & 0 & -1 \\
\end{bmatrix}
\begin{bmatrix}
\Delta a_{A} \\
 b_{A} \\
 c_{A} \\
 \Delta d_{A} \\
 u_{A} \\
 v_{A}
\end{bmatrix}.
$}
\label{equ:threeOffsets}
\end{equation}

The above six geometric parameters of affine transformations can be linearly mapped to the positional offsets of three corners. This linear relationship enables the effective estimation of affine transformation parameters using neural networks, in a manner similar to the estimation of positional offsets.

\subsection{Geometric Parameterization from Degenerate Homography in SKS Decomposition}

\begin{figure}[t]
\begin{center}
\includegraphics[width=0.95\linewidth]{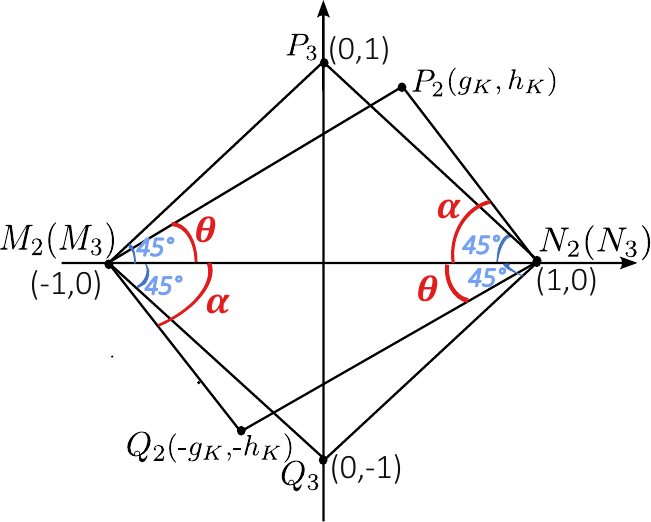}
\end{center}
\vspace{-3mm}
\caption{Degenerate geometric interpretation of the kernel transformation in an affine transformation. Specifically, the four distinct angular offsets described in the main manuscript are reduced to two identical angular offsets, $\theta$ and $\alpha$, under the affine transformation.}
\label{fig:Geo_mean_HK_Affine}
\vspace{-2mm}
\end{figure}

At the same time, geometric parameterization of affine transformations can also be derived from the simplification to homography, since affine transformation is a degenerate case of homography. As shown in \cref{fig:Geo_mean_HK_Affine}, the four angles \{$\theta$, $\alpha$, $\beta$, $\gamma$\} in Fig.~\textcolor{red}{4} of the main manuscript under an affine transformation satisfy: $\theta=\gamma$ and $\alpha=\beta$. Consequently, the 6-DOF affine distortion, which maps a square to a parallelogram, can be decoupled into four positional offsets and two angular offsets. The kernel transformation, $\mathbf{H}_K$, under an affine transformation is reduced to 2-DOF and is expressed as:
\begin{equation}
\resizebox{0.86\linewidth}{!}{$
\mathbf{H}_K^{(a\!f\!f.)} = \begin{bmatrix}
\Delta a_{K}\!+\!1 & u_{K}  & \\
       &  1     &  \\
   &  & \Delta a_{K}\!+\!1
\end{bmatrix} = \begin{bmatrix}
1 & g_{K}  &  \\
   &  h_{K}     &  \\
  &  & 1
\end{bmatrix}.
$}
\label{equ:HK_expression_affine}
\end{equation}

Referring to Eq.~(\textcolor{red}{11}) of the main manuscript, the entire affine transformation $\bar{\mathbf{H}}_A$ is decomposed into:
\begin{equation}
    \bar{\mathbf{H}}_A =  \mathbf{H}_{T}^{-1} {\mathbf{H}}_S  \mathbf{H}_{T} {\mathbf{H}}_{S_2}^{-1} {\mathbf{H}}_{K}^{(a\!f\!f.)}{\mathbf{H}}_{S_2}.
\end{equation}

Combing \cref{equ:H_A_bar} and substituting \cref{equ:HK_expression_affine} into the above equation, ${\mathbf{H}}_A$ will be expressed as 
\begin{equation}
\resizebox{0.86\linewidth}{!}{$
\begin{split}
\mathbf{H}_A =& \begin{bmatrix}
\Delta a_{S}\!+\!1 & -b_{S}  & u_{S} \\
b_{S}   & \Delta a_{S}\!+\!1 & v_{S} \\
      &      & 1 
\end{bmatrix}*\begin{bmatrix}
r & r  &  \\
-r   & r &  \\
       &   & 1 
\end{bmatrix} * \\
& \begin{bmatrix}
1 & g_{K}  &  \\
   &  h_{K}     &  \\
  &  & 1
\end{bmatrix}*
\begin{bmatrix}
r & r  &  \\
-r   & r &  \\
       &   & 1 
\end{bmatrix}^{-1},
\end{split}
$}
\label{equ:HA_expression_SKS}
\end{equation}
with the six effective elements expressed by
\begin{equation}
\resizebox{0.86\linewidth}{!}{$
\begin{split}
\Delta a_A  &= \frac{(\Delta a_{S}\!+\!1)(g_K+h_K+1)+b_{S}(g_K-h_K+1)}{2} -1, \\
b_A &= \frac{(\Delta a_{S}\!+\!1)(g_K+h_K-1)+b_{S}(g_K-h_K-1)}{2},  \\
c_A &= \frac{(\Delta a_{S}\!+\!1)(h_K-g_K-1)+b_{S}(g_K+h_K+1)}{2},  \\
\Delta d_A &= \frac{(\Delta a_{S}\!+\!1)(h_K-g_K+1)+b_{S}(g_K+h_K-1)}{2} -1, \\
u_A &= u_S, \\
v_A &= v_S. \\
\end{split}
$}
\label{equ:paras_expression_affine}
\end{equation}

Considering the linear relationships illustrated in \cref{equ:threeOffsets}, the three-corner positional offsets can be expressed as quadratic polynomials in the six geometric parameters \{$\Delta a_S$, $b_S$, $u_S$, $v_S$, $g_K$, $h_K$\}. This form of geometric parameterization for affine transformations is also well-suited for neural network fitting, as supported by the universal approximation theory~\cite{Hornik1989UniversalApproximators}.

\section{Details for Unified Geometric Parameterization of 2D Transformations}
\label{sec:affine}

Based on the analysis of similarity transformations in the main manuscript and the affine transformations discussed above, a unified solution is proposed to estimate 2D transformations using eight geometric parameters.

\subsection{Estimation of 6-DOF Affine Transformations} 
For image pairs undergoing affine transformations, when processed by a neural network trained for homography estimation, eight geometric parameters are still predicted. Among these parameters, two explicitly indicate whether the estimated homography degenerates to an affine transformation. As illustrated in Fig.~\textcolor{red}{1} of the main manuscript, the hallmark of an affine transformation is its ability to map a square in the source image to a parallelogram in the target image.

In practice, the criterion for affine transformations shown in the main manuscript is implemented as:
\begin{equation}
    \max(b_K, v_K) < thresh_1,
    \label{equ:affine_th1}
\end{equation}
where $thresh_1$ implicitly constrains the angular difference between the two pairs of opposite angles.

\subsection{Estimation of 4-DOF Similarity Transformations} 
Moreover, the criterion for similarity transformations in the main manuscript is implemented as:
\begin{equation}
    \max(b_K, v_K, \Delta a_K, u_K) < thresh_2,
    \label{equ:similarity_th2}
\end{equation}
where $thresh_2$ implicitly constrains the angular deviation of all four angles from $45^\circ$ (referring to Eq.~\textcolor{red}{17-20} of the main manuscript).


\section{Additional Experiments}
\vspace{-1mm}

\subsection{Detailed Configurations}
\vspace{-1mm}

\noindent \textbf{Data generation.} We evaluate the unified homography parameterization across three distinct datasets: synthetic MSCOCO~\cite{MSCOCO}, dynamic SPID~\cite{SPID}, and cross-modal GoogleMap~\cite{CLKN}. The generation of similarity and homography image pairs has been introduced in the main manuscript. For affine image pairs, random perturbations ranging from -32 to 32 are applied to the four corner points of the input images, except for the lower-left corner.

\vspace{1mm}
\noindent \textbf{Network configuration.} Tests are conducted across several homography estimation networks, including the pioneering DHN~\cite{DHN16}, the recent state-of-the-art (SOTA) RHWF~\cite{RHWF}, and MCNet~\cite{MCNet}. We adhere to all default training settings as specified in the respective implementations of these deep homography methods, ensuring a fair comparison.

\subsection{Deep Similarity and Affine Estimation with Trained Homography Networks}
\vspace{-1mm}

In addition to the results in Sec.~\textcolor{red}{4.5} of the main manuscript, we further evaluate the aforementioned datasets using the following two additional metrics.

\begin{figure*}[t]
    \centering
    \begin{subfigure}{0.32\textwidth}
        \centering
        \includegraphics[width=\textwidth]{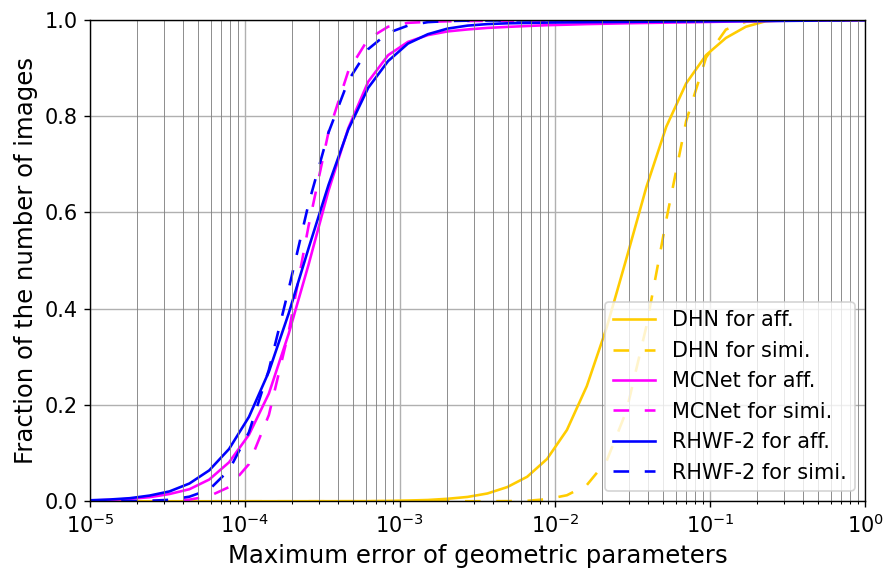}
         \vspace{-3mm}
        \caption{MSCOCO}
    \end{subfigure} \,
    \begin{subfigure}{0.32\textwidth}
        \centering
        \includegraphics[width=\textwidth]{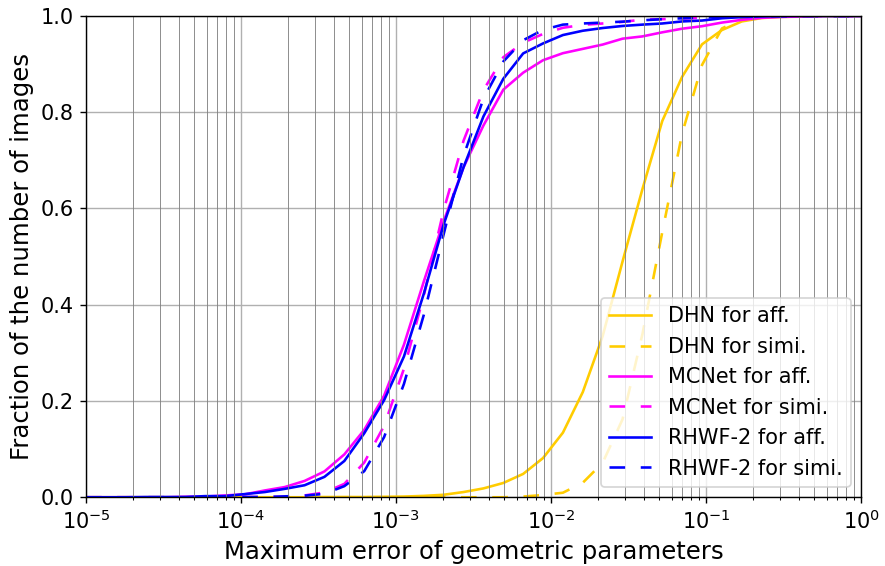}
         \vspace{-3mm}
        \caption{SPID}
    \end{subfigure} \,
    \begin{subfigure}{0.32\textwidth}
        \centering
        \includegraphics[width=\textwidth]{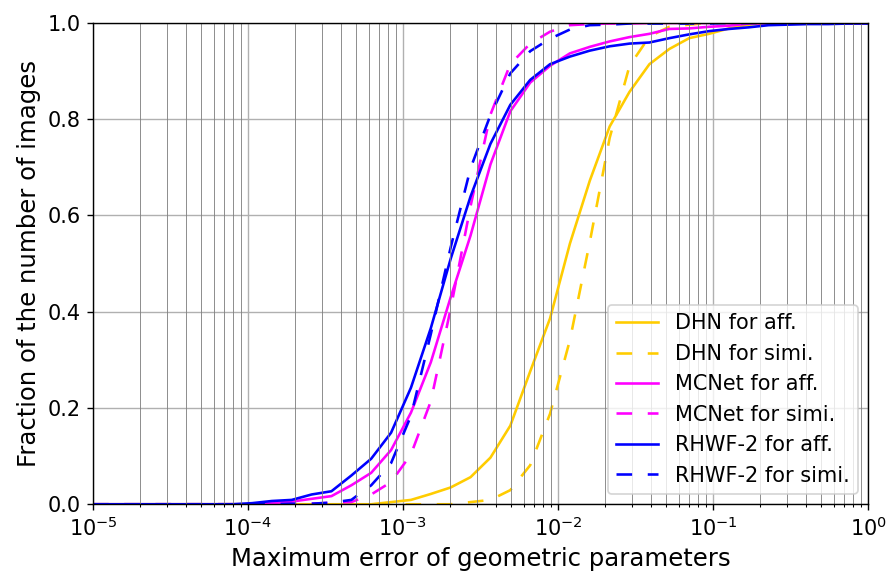}
         \vspace{-3mm}
        \caption{GoogleMap}
    \end{subfigure}
    \vspace{-2mm}
    \caption{Maximum error of geometric parameters for affine and similarity estimation across three distinct datasets. Among the proposed eight geometric parameters of homography, the maximum error of \{$b_K$, $v_K$\} and \{$b_K$, $v_K$, $\Delta a_K$, $u_K$\} is used to evaluate  affine and similarity, respectively.}
    \label{fig:Maxresults}
\end{figure*}

\vspace{1mm}
\noindent \textbf{Evaluation with maximum error of geometric parameters.} This metric directly evaluates the criteria for 6-DOF affine transformations and 4-DOF similarity transformations, as described in \cref{equ:affine_th1} and \cref{equ:similarity_th2}, respectively. The proposed homography parameterization is evaluated across the three datasets and three network architectures. As illustrated in~\cref{fig:Maxresults}, the overall results indicate that the eight geometric parameters perform similarly for both affine and similarity transformations. The maximum error of geometric parameters under similarity transformations is slightly lower than that under affine transformations.

\begin{figure*}[t]
    \centering
    \begin{subfigure}{0.32\textwidth}
        \centering
        \includegraphics[width=\textwidth]{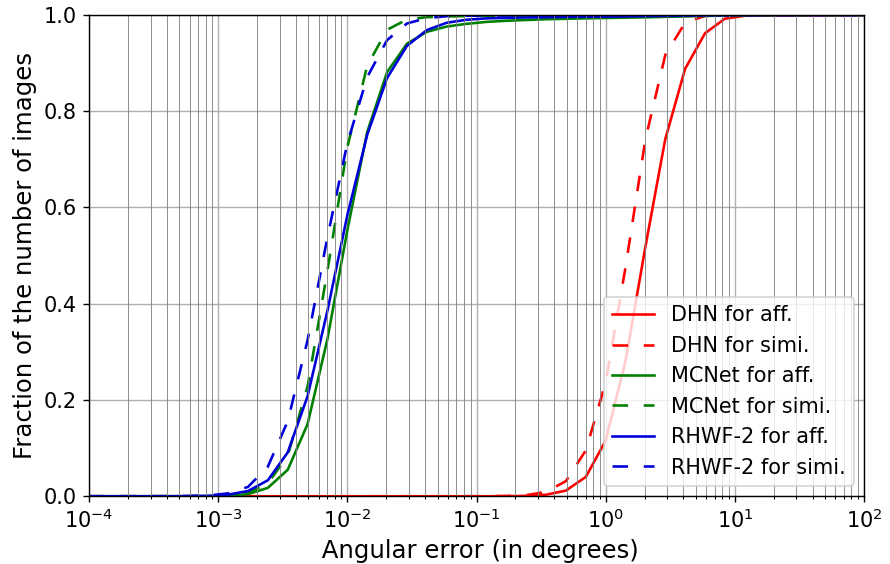}
         \vspace{-3mm}
        \caption{MSCOCO}
    \end{subfigure} \,
    \begin{subfigure}{0.32\textwidth}
        \centering
        \includegraphics[width=\textwidth]{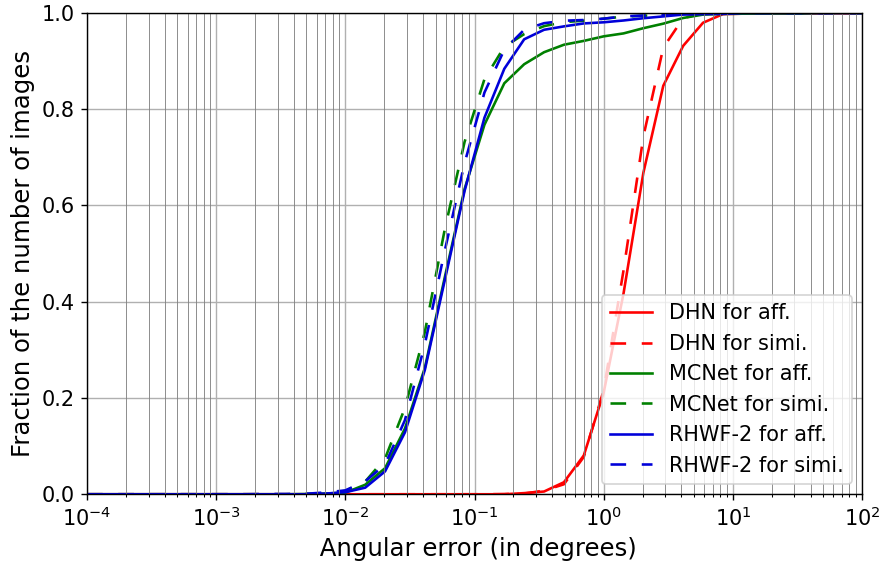}
         \vspace{-3mm}
        \caption{SPID}
    \end{subfigure} \,
    \begin{subfigure}{0.32\textwidth}
        \centering
        \includegraphics[width=\textwidth]{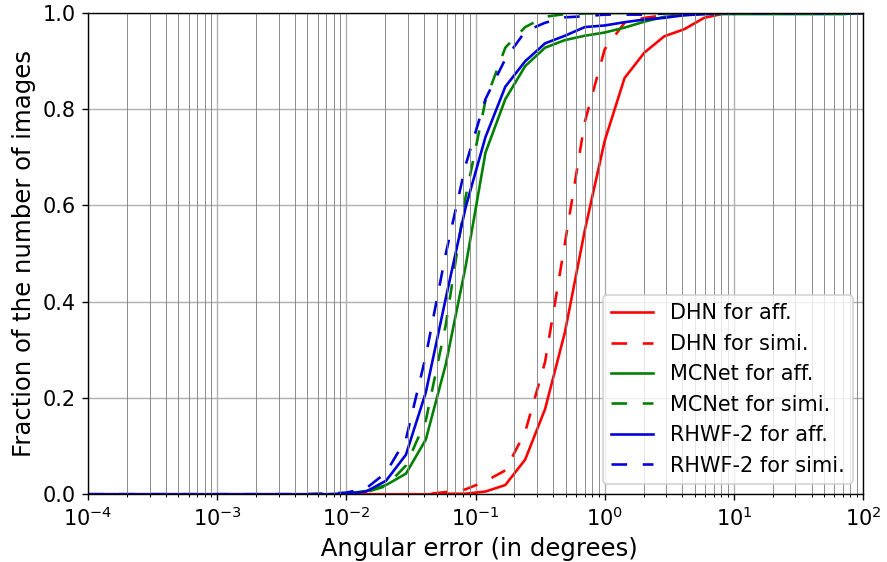}
         \vspace{-3mm}
        \caption{GoogleMap}
    \end{subfigure}
    \vspace{-2mm}
    \caption{Angular error for affine and similarity estimation across three distinct datasets. The average angular difference between the two pairs of opposite angles is used to evaluate similarity, while the average angular difference of the four angles from $45^{\circ}$ is used to evaluate affine transformations.}
    \label{fig:Angleresults}
\end{figure*}

\vspace{1mm}
\noindent \textbf{Evaluation with average error of two pairs of opposite angles and four angles from $45^\circ$.} This metric evaluates the unified geometric parameterization of 2D transformations from a different perspective. The homography parameterization is again evaluated across three datasets and three network architectures, as shown in \cref{fig:Angleresults}. The average error of the two pairs of opposite angles and the four angles from $45^\circ$ is used to evaluate similarity and affine transformations, respectively. Both of them show strong consistency with the maximum error of geometric parameters, suggesting that the metric is satisfactory for practical use due to its negligible computational cost.

From the above experiments, a clear correspondence is observed between the maximum error of geometric parameters and the average angular errors. For example, for MCNet and RHWF-2, a maximum error of geometric parameters of $0.01$ corresponds to an angular deviation of approximately $0.3^\circ$ for both pairs of opposite angles, as well as all four angles from $45^\circ$.

\end{document}